%% file: main.tex
\pdfoutput=1

\documentclass[11pt]{article}

\usepackage[]{EMNLP2022}

\usepackage{times}
\usepackage{latexsym}

\usepackage[T1]{fontenc}

\usepackage[utf8]{inputenc}

\usepackage{microtype}

\usepackage{amsmath}
\usepackage{amsfonts}
\usepackage{amssymb}
\usepackage{amsthm}
\usepackage{booktabs}

\usepackage{graphicx}
\usepackage{booktabs}
\usepackage{bm}
\usepackage{hyperref}
\usepackage{xspace}
\usepackage{makecell}
\usepackage{multirow}

\usepackage{subfigure}
\usepackage{enumitem}
\usepackage{url}
\usepackage[flushleft]{threeparttable}
\usepackage{longtable}
\usepackage{supertabular}
\usepackage{microtype}
\usepackage[normalem]{ulem}

\RequirePackage{xspace}
\usepackage{nicefrac}       
\usepackage{microtype}      
\usepackage{lipsum}
\usepackage{natbib}

\usepackage{flushend}
\usepackage{enumerate}
\usepackage{subfigure}
\usepackage[T1]{fontenc}
\usepackage{framed}
\usepackage{stmaryrd}
\usepackage{enumerate}
\usepackage{enumitem}
\usepackage{graphicx}
\usepackage{hyperref}
\usepackage{wrapfig}
\usepackage{epstopdf}
\usepackage{url}
\usepackage[T1]{fontenc}
\usepackage{listings}
\usepackage{flushend}

\usepackage{cleveref}
\usepackage{xargs}
\usepackage[colorinlistoftodos,prependcaption,textsize=tiny]{todonotes}
\usepackage[group-separator={,}, group-minimum-digits=4]{siunitx}

\newcommandx{\yli}[2][1=]{\todo[linecolor=blue,backgroundcolor=blue!25,bordercolor=blue,#1]{#2}}
\newcommandx{\improvement}[2][1=]{\todo[linecolor=yellow,backgroundcolor=yellow!25,bordercolor=yellow,#1]{#2}}
\newcommandx{\thiswillnotshow}[2][1=]{\todo[disable,#1]{#2}}

\crefname{section}{§}{§§}
\Crefname{section}{§}{§§}

\newcommand{\ours}{\textsc{ReSel}\xspace}

\newcommand{\etc}{\emph{etc.}\xspace} 
\newcommand{\ie}{\emph{i.e.}\xspace} 
\newcommand{\eg}{\emph{e.g.}\xspace} 

\newcommand{\editout}[1]{}

\newcommand{\chao}[1]{}

\title{ReSel: $N$-ary Relation Extraction from Scientific Text and Tables by\\
  Learning to Retrieve and Select
}

\author{Yuchen Zhuang$^1$, Yinghao Li$^1$, Jerry Junyang Cheung$^1$, Yue Yu$^1$,\\ {\bf Yingjun Mou$^1$, Xiang Chen$^2$, Le Song$^{3,4}$, Chao Zhang$^1$}\\
$^1$Georgia Institute of Technology, Atlanta, USA\\
$^2$Adobe Research, San Jose, USA\quad
$^3$MBZUAI, Abu Dhabi, UAE\quad $^4$BioMap, Beijing, China\\
$^1$\texttt{$\{$yczhuang,yinghaoli,jzhang3027,yueyu,ymou32,}\\
\texttt{chaozhang$\}$@gatech.edu}\\
$^2$\texttt{xiangche@adobe.com}\quad $^3$\texttt{le.song@mbzuai.ac.ae} 
}

\begin{document}
\maketitle

\input{0abstract}

\input{1introduction}

\input{1half-related}

\input{2problem}

\input{3method}
\input{4experiments}

\input{5conclusion}

\input{6limit}

\input{ack}
\bibliography{anthology,custom}
\bibliographystyle{acl_natbib}

\clearpage

\appendix
\input{appendices/appendix}
\input{appendices/appendix2}
\input{appendices/appendixT}
\input{appendices/appendix3}
\input{appendices/appendix4}
\input{appendices/appendix5}

\end{document}

%% file: 0abstract.tex
\begin{abstract}
We study the problem of extracting $N$-ary relation tuples from scientific articles.
This task is challenging because the target knowledge tuples can reside in multiple parts and modalities of the document.
Our proposed method \ours decomposes this task into a two-stage procedure that first \emph{retrieves} the most relevant paragraph/table and then \emph{selects} the target entity from the retrieved component.
For the high-level retrieval stage, \ours designs a simple and effective feature set, which captures multi-level lexical and semantic similarities between the query and components.
For the low-level selection stage, \ours designs a cross-modal entity correlation graph along with  a multi-view architecture, which models both semantic and document-structural relations between entities.
Our experiments on three scientific information extraction datasets show that \ours outperforms state-of-the-art baselines significantly.~\footnote{Our code is available on \url{https://github.com/night-chen/ReSel}.}
\end{abstract}

%% file: 1introduction.tex
\section{Introduction}\label{sec:intro}

\begin{figure}[t]
  \centering
  \includegraphics[width=\linewidth]{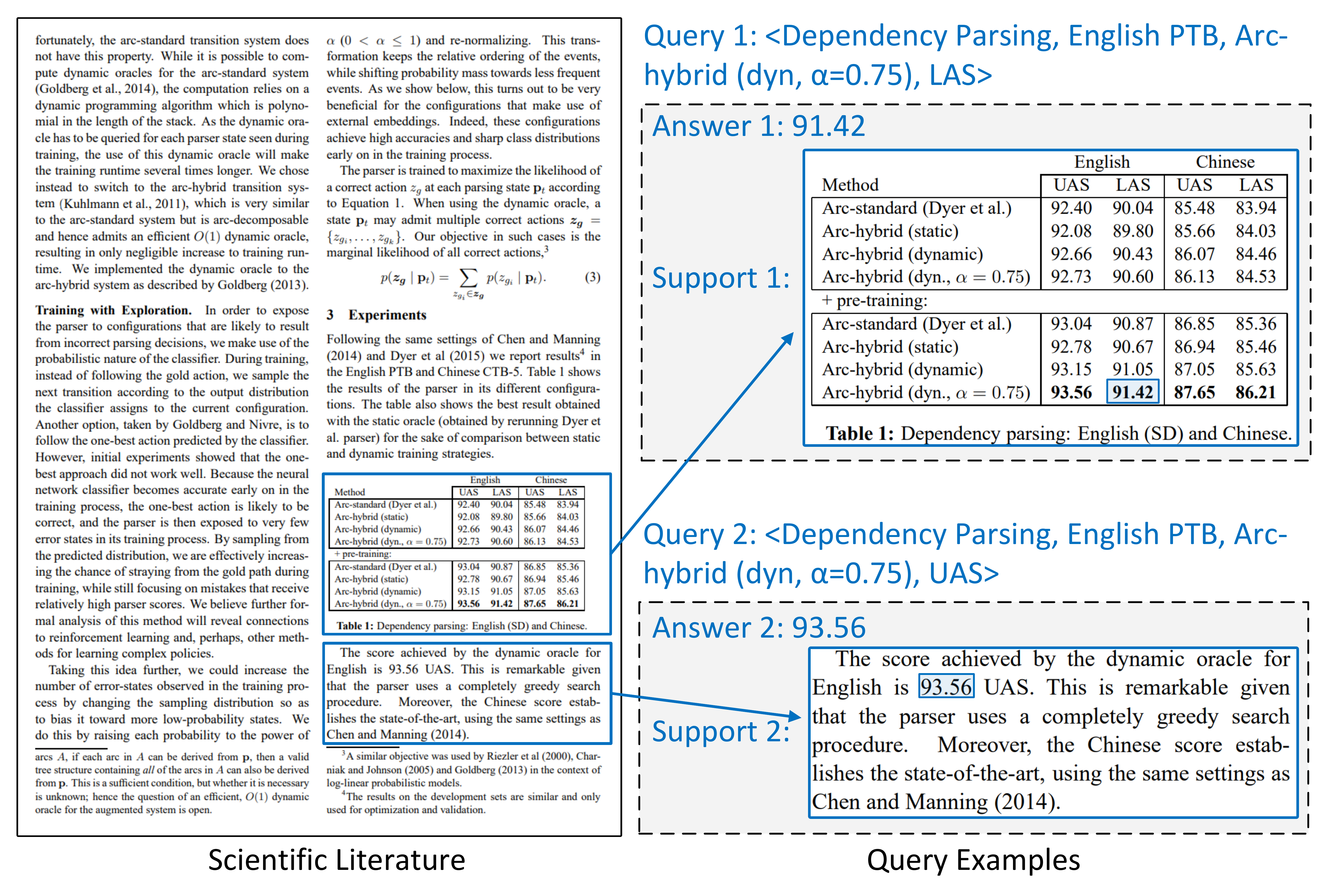}
  \vspace{-3ex}
  \caption{
    Illustration of the multi-modal scientific $N$-ary relation extraction problem on
    the SciREX dataset.}
  \vspace{-1.5ex}
  \label{fig:intro-example}
\end{figure}

Scientific information extraction (SciIE) \cite{augenstein2017semeval, luan2018multi, jiang2019role}, the task of extracting scientific concepts along with their relations from scientific literature corpora, is important for researchers to keep abreast of latest scientific advances.
A key subtask of SciIE is the $N$-ary relation extraction problem~\cite{jia2019document, jain2020scirex}, which aims to extract the relations of different entities as $N$-ary knowledge tuples.
This problem is challenging because the entities of the knowledge tuples often reside in multiple sections (\eg, abstracts, experiments) and modalities (\eg, paragraphs, tables, figures) of the document.
Effective scientific $N$-ary relation extraction requires not only understanding the semantics of different modalities, but also performing document-level inference based on interleaving signals such as co-occurrences, co-references, and structural relations, as shown in Figure \ref{fig:intro-example}.

Document-level
$N$-ary relation extraction  has been studied in literature~\cite{jia2019document, jain2020scirex, viswanathan2021citationie, liu2021role}. Some  works \cite{zeng2020double, tu2019multi} use graph-based approaches to model long-distance relations in the document with the focus on text only.
However, for scientific articles, an equally if not more important data structure is the table, as scientific results are often reported in tables and then referred to and discussed in text.
There are also works that pre-train large-scale transformer models on massive table and text pairs~\cite{yin2020tabert, herzig2020tapas}.
These methods are designed for question answering, which are strong at retrieving answers that semantically match the query but fall short in inferring fine-grained entity-level $N$-ary relations.
Besides, to perform well on SciIE, they usually require large task-specific data to fine-tune the pre-trained model, especially for long documents that contain many candidates.
But in practice, such large-scale annotation data can be expensive and  labor-intensive to curate.
Therefore, extracting $N$-ary relations jointly from scientific text and tables still remains an important but challenging problem.


We propose \ours, a hierarchical \textbf{Re}trieve-and-\textbf{Sel}ection model for multi-modal and document-level SciIE.
In \ours, we pose the $N$-ary relation extraction problem as a question answering task over text and tables (Figure \ref{fig:intro-example}). \ours then decomposes the challenging task into two simpler sub-tasks: (1) high-level component retrieval, which aims to locate the target paragraph/table where the final target entity resides, and (2) low-level entity extraction, which aims to select the target entity from the chosen component.


For high-level component (\emph{i.e.}, paragraph or table) retrieval, we design a feature set that combines the strengths of two classes of retrieval methods: (1) \emph{sparse retrieval}~\cite{aizawa2003information, robertson2009probabilistic} that represents the query-candidate pairs as high-dimensional sparse vectors to encode lexical features; (2) \emph{dense retrieval}~\cite{karpukhin2020dense} that leverages latent semantic embeddings to represent query and candidates.
We design sparse and dense retrieval features for query-component pairs by augmenting BERT~\cite{devlin-etal-2019-bert}-based semantic similarities with entity-level semantic and lexical similarities, allowing for training an accurate high-level retriever using only a small amount of labeled data.



The low-level entity extraction stage aims to infer $N$-ary entity relations from complex and noisy signals across paragraphs and tables.
In this stage, we first build a cross-modal entity-correlation graph, which  encodes different entity-entity relations such as co-occurrence, co-reference, and table structural relations.
While most of the existing methods~\cite{zheng2020document, zeng2020double} use BERT embeddings as node representations, we find BERT embeddings limited in distinguishing adjacent table cells or similar entities.
This issue is even more severe when the BERT embeddings are propagated on the graph.
To address this, we design a new \emph{bag-of-neighbors (BON)} representation.
It computes the lexical and semantic similarities between each candidate entity and its 1-hop neighbors.
We then feed the BON features into a graph attention network (GAT) to capture both neighboring semantics and structural correlations.
Such GAT-learned features and BERT-based embeddings are treated as two complementary views, which are co-trained with a consistency loss.


We summarize our key contributions as follows:
(1) We propose a hierarchical retrieve-and-select learning method that decomposes $N$-ary scientific relation extraction into two simpler subtasks;
(2) For high-level component retrieval, we propose a simple but effective feature-based model that combines multi-level semantic and lexical features between queries and components;
(3) For low-level entity extraction, we propose a multi-view architecture, which  fuses graph-based structural relations with BERT-based semantic information for extraction;
(4) Extensive experiments on three datasets  show the superiority of 
both the high-level and low-level modules in \ours.

%% file: 1half-related.tex
\section{Related Work}\label{sec:related}
\paragraph{Component Retrieval}
For component retrieval, traditional sparse retrieval methods such as 
TF-IDF~\cite{aizawa2003information} and 
BM25~\cite{robertson2009probabilistic} focus on keyword-level matching but ignore entity semantics.
Recently, pre-trained language models have also been used to  represent queries and documents in a learned space~\cite{karpukhin2020dense} and have been extended to handle tabular context~\cite{herzig-etal-2021-open,ma-etal-2022-open}. 
However, these methods mainly focus on passage-level retrieval, and cannot well  capture fine-grained entity-level semantics~\cite{zhang2020revisiting, su2021whitening}. 
Such an issue makes them suboptimal for encoding nuanced terms and descriptions in scientific articles.
In contrast, \ours leverages both component- and entity-level semantic and lexical features that help the model better understand the correlations between components and queries.

\paragraph{$N$-ary Relation Extraction} 
Many existing methods~\cite{jia2019document, jain2020scirex, viswanathan2021citationie} treat $N$-ary relation extraction as a binary classification problem and predict whether the composition of $N$ entities in the document are valid or not.
However, the candidate space grows exponentially with $N$, and the performance of the binary classifiers can be largely influenced by the number and quality of negative tuples.
Some other methods~\cite{du-etal-2021-grit, huang-etal-2021-document} formulate the problem as role-filler entity extraction and propose BERT-based generative models to extract the correct entities for each element of the $N$-ary relation.
None of these methods consider $N$-ary relation across modalities. 
\citet{lockard-etal-2020-zeroshotceres}  leverages the layout information for  extracting relations from web pages.
However, the layout information in science articles are less prominent and harder to be utilized.

%% file: 2problem.tex
\begin{figure*}[t]
\centering
\includegraphics[width=\linewidth]{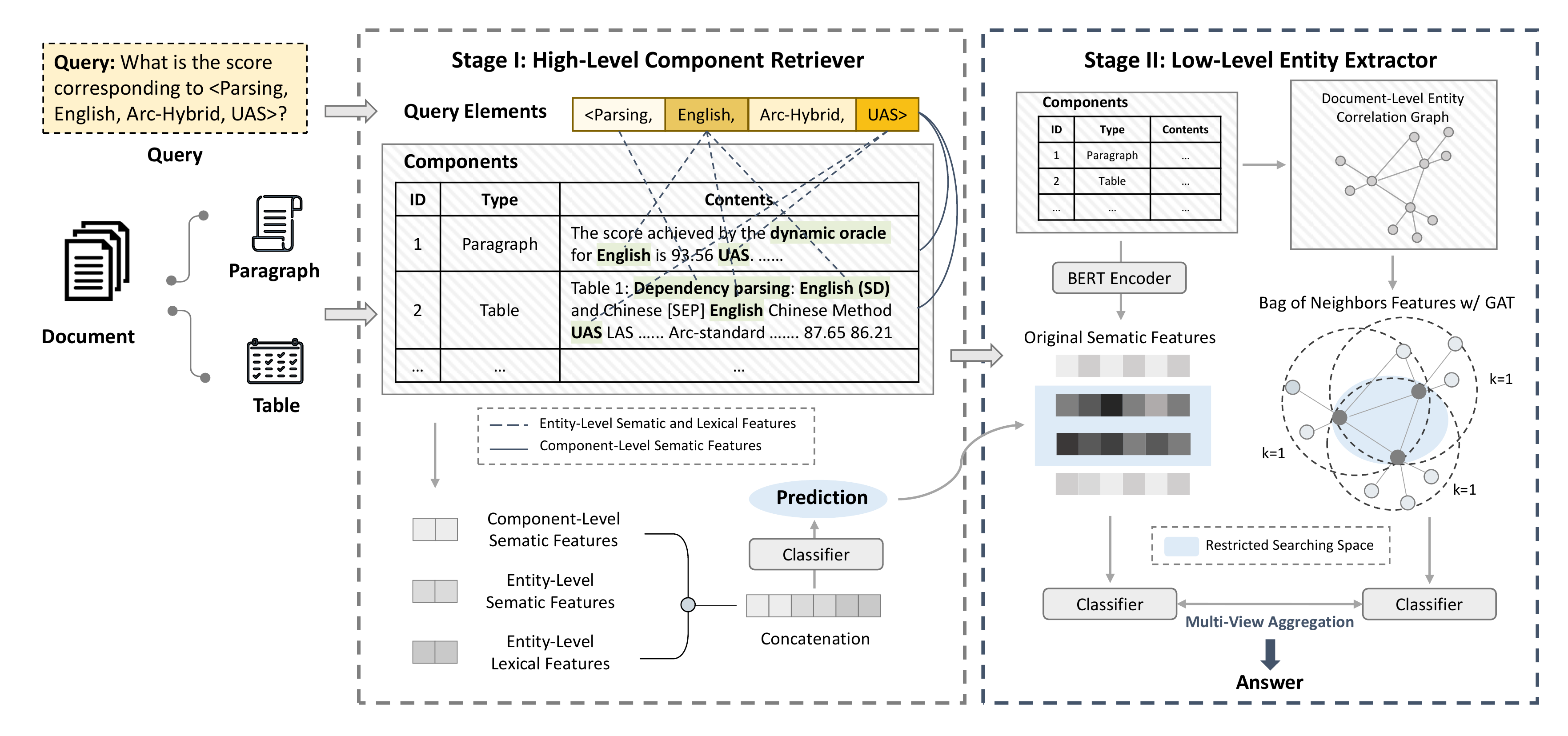}
\vspace{-3ex}
\caption{Illustration of \ours two-stage pipeline: high-level component retrieval and low-level entity extractor.}
\label{fig:mainmethod}
\end{figure*}

\vspace{-0.6ex}
\section{Problem Formulation}\label{sec:pf}
\vspace{-0.6ex}
In SciIE, we
aim to extract information from a corpus of $M$ scientific articles.
Each article, denoted as $\mathcal{D}$, is a sequence of $|\mathcal{D}|$ components, where each component $C_i\in\mathcal{D}$, $i\in\mathbb{N}_{[0,|\mathcal{D}|)}$ can be either a paragraph or a table.
A table is flattened and concatenated with its caption as a sequence of words.
Given a document $\mathcal{D}$, we have a set of queries $\mathcal{Q}$ with number $|\mathcal{Q}|$.
Each query contains $N-1$ elements $Q_j=[e_{j,1},\cdots,e_{j,N-1}],\ j\in\mathbb{N}_{[0,|\mathcal{Q}|)}$, and the
task is to extract the correct $N$-th element from document $\mathcal{D}$ to form a valid $N$-ary relation.
We assume a dataset  $\{x_k,y_k\}_{k=1}^M$
that can be used to learn such a
$N$-ary relation extractor,
each sample includes a document and a set of queries
$x_{k}=(\mathcal{D}_k,\mathcal{Q}_k)$, and each ground-truth label $y_k$
indicates the target entity in the document $\mathcal{D}_k$.


%% file: 3method.tex
\vspace{-0.6ex}
\section{ The \ours Method}
\vspace{-0.6ex}
\label{sec:method}

\input{30overview}
\input{31high}
\input{32low}

\input{33trainobj}

%% file: 31high.tex
\subsection{Component Retriever}
\label{subsec:method2}



In Stage I, we design 
a high-level model to retrieve the most relevant paragraphs or tables that contain the final answer. 
We first use BERT to embed the paragraphs/tables into sequences of vectors (details in Appendix~\ref{appendix:ce}).
We encode the $j$-th query $Q_j=[e_{j,1},\cdots,e_{j,N-1}]$, into query embedding $\mathbf{h}(Q_j)$ and get the corresponding element embeddings of the query $\mathbf{h}(e_{j,a}),\ a\in\mathbb{N}_{[0,N)}$.
Similar to the query encoder, we encode the $i$-th component, $C_i$, as component embedding $\mathbf{h}({C_i})$, and the
averaged entity embeddings $\mathbf{h}({m_{i,b}})$, where $m_{i,b}\in C_i$, indicates the $b$-th entity extracted from $C_i$.
With the encoded sequences of vectors, we compute the different views of features for the component-query pair $(C_i,Q_j)$ as follows to
take advantage of both the entity-level matching signals and component-level semantic signals, which are complementary:

\textbf{Component-Level Semantic Features (CS).}
The first view extracts the semantic features for component-query pairs from two different angles:
(1) \textit{Embedding-Based Similarity:}
the cosine similarities $f_{\operatorname{cs-1}}(C_i,Q_j)$ between component and query embeddings.
(2) \textit{Entailment-Based Score:}
the classification score $f_{\operatorname{cs-2}}(C_i,Q_j)$ between $Q_j$ and $C_i$ calculated by feeding them both into a BERT binary sequence classifier as a concatenated sequence~\citep{nogueira2019passage, nie2019revealing}.
We concat these two scalar features as the first view $\mathbf{f}_{\operatorname{cs}}(C_i,Q_j)$.


\textbf{Entity-Level Semantic Features (ES).} The second view computes \emph{entity-level} cosine similarities $f_{\operatorname{es}}(m_{i,b},e_{j,a})$ between the component entity embeddings $\mathbf{h}({m_{i,b}})$ and the query elements embeddings $\mathbf{h}({e_{j,a}})$.
With all these similarity scores, we apply a max-pooling operation over all
component entities $m_{i,b}$, and use the obtained maximum $f_{\operatorname{es}}(C_i,e_{j,a})=\max_{m_{i,b}\in C_i}f_{\operatorname{es}}(m_{i,b},e_{j,a})$ to represent the relation between the component $C_i$ and one query element $e_{j,a}$.
Then, we gather the relation scores $f_{\operatorname{es}}(C_i,e_{j,a})$ as the final entity-level semantic feature vector:
$\mathbf{f}_{\operatorname{es}}(C_i,Q_j)=[f_{\operatorname{es}}(C_i,e_{\operatorname{j,1}}), \cdots, f_{\operatorname{es}}(C_i,e_{\operatorname{j,N-1}})]^{\sf{T}}$.


\textbf{Entity-Level Lexical Features (EL).} Our third view extracts lexical features between component entities and the query elements. 
We compute three text similarities (Appendix~\ref{appendix:textsim}):
(1) \textit{Levenshtein Distance}~\cite{levenshtein1966binary}; 
(2) the length of \textit{Longest Common Substring}; 
(3) the length of \textit{Longest Common Subsequence}. 
As the metrics vary in scale according to the length of the strings, we use the normalized metrics $\mathbf{f}_{\operatorname{el}}(m_{i,b},e_{j,a})\in[0,1]^3$ via dividing by involved string lengths.
Similar to ES features, we perform max-pooling to obtain the relation scores between the component and a single query element, $\mathbf{f}_{\operatorname{el}}(C_i,e_{j,a})=\max_{m_{i,b}\in C_i}\mathbf{f}_{\operatorname{el}}(m_{i,b},e_{j,a})$ and concatenate the results as entity-level lexical features:
$\mathbf{f}_{\operatorname{el}}(C_i,Q_j)=[\mathbf{f}_{\operatorname{el}}(C_i,e_{j,1})^{\sf{T}}\oplus \dots \oplus \mathbf{f}_{\operatorname{el}}(C_i,e_{j,N-1})^{\sf{T}}]^{\sf{T}}$.

We aggregate the features to predict which component has the highest probability to contain the final answer.
As the features in the three views share the same scale range and similar dimensionality, we just concatenate these features together as $\mathbf{f}^{\operatorname{h}}=[\mathbf{f}_{\operatorname{cs}}^{\sf{T}}\oplus\mathbf{f}_{\operatorname{es}}^{\sf{T}}\oplus\mathbf{f}_{\operatorname{el}}^{\sf{T}}]^{\sf{T}}$,
and train one unified classifier over $\mathbf{f}^{\operatorname{h}}$ for component retrieval.

%% file: 32low.tex
\subsection{Entity Extractor}
\label{subsec:method3}

In Stage II, we use
the predictions from Stage I to restrict the searching space for low-level
entity extraction.
\subsubsection{Multi-Modal Entity-Level Graph}\label{subsubsec:method41}
\begin{figure}[t]
\centering
\includegraphics[width=0.9\linewidth]{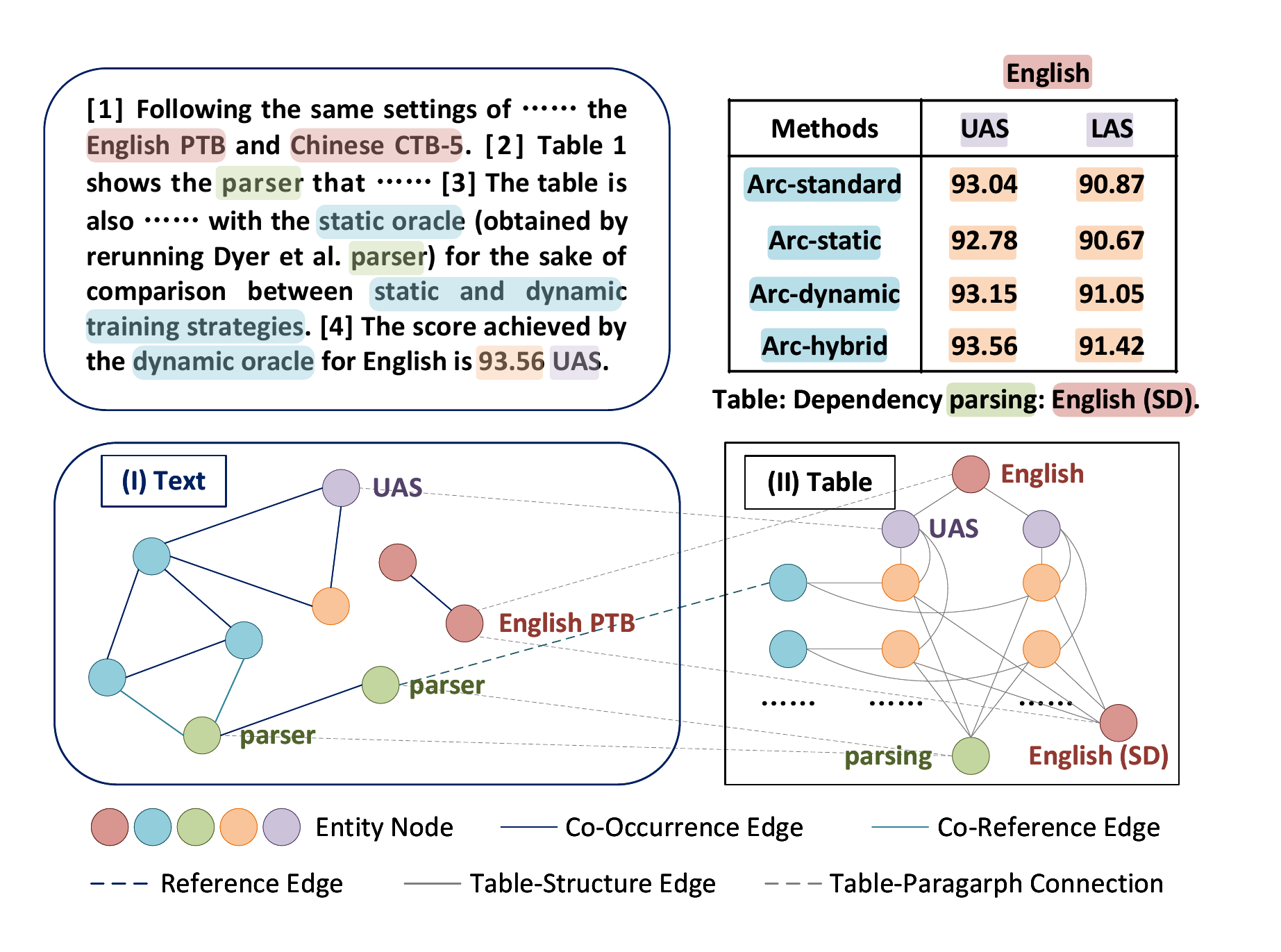}
\caption{Illustration of constructing the multi-modal entity correlation graph between paragraphs and tables.}
\label{fig:low-graphexample}
\end{figure}

To model document-level entity
correlations, we construct a multi-modal entity correlation graph $\mathcal{G}=(\mathcal{V},\mathcal{E})$, where $\mathcal{V}=\{v_1,v_2,\cdots,v_{|\mathcal{V}|}\}$ 
denotes the entity nodes,
and
$\mathcal{E}=\{E_{(v_i,v_j)}|v_i,v_j\in\mathcal{V};i,j\in\mathbb{N}_{[1,|\mathcal{V}|]}\}$
denotes the edges between them.
Each node $v_i\in\mathcal{V}$ represents a paragraph entity or a table cell.
 We construct different edge types to model
the intra- and inter-modality relations to encode the entity
correlation across modalities as in Figure \ref{fig:low-graphexample}:
(1) \textbf{Co-occurence Edge} measures whether two entity nodes $v_i$ and $v_j$ occur in the same sentence or adjacent sentences; 
(2) \textbf{Co-reference Edge} extracts the relation information of two entity nodes $v_i$ and $v_j$ referring to the same concept; 
(3) \textbf{Reference Edge} bridges the table and text with reference information (\eg, ``in Table 3''); 
(4) \textbf{Table-Structure Edge} extracts the structural information of columns and rows of tables; 
(5) \textbf{Table-Paragraph Connection} enhances the linking between table cells and paragraph entities via text similarities (detailed in Appendix~\ref{appendix:graph}).

With these five edge types from different modalities covering nearly all hidden relations in the document, the multi-modal entity correlation graph can effectively model document-level information.
As all edge types are ranged in $[0,1]$ and most of them do not overlap, we treat them equally and define the graph as an undirected homogeneous graph. 

\subsubsection{Bag-of-Neighbors Features}\label{subsubsec:method42}

For low-level entity extraction from the retrieved paragraph/table, a key challenge is that the entities (nodes) in the same sentence or adjacent table cells can have very similar BERT embeddings and hard to be discriminated by a BERT-only classifier.
Further, such entities often share many common neighbors on the graph, which means their embeddings can be easily further over-smoothed when propagated on the graph.
To tackle these challenges, we propose the bag-of-neighbors (BON) features (Figure~\ref{fig:bon}) based on the entity-level semantic and lexical features.
Given an entity node $v_i$ and a query $Q_j$, we define the initial embeddings as:
\begin{equation}
    \begin{aligned}
    \mathbf{g}(v_i)=[&f_{\operatorname{es}}(v_i,e_{j,1})\oplus\cdots\oplus f_{\operatorname{es}}(v_i,e_{j,N-1})\oplus\\
    &\mathbf{f}_{\operatorname{el}}(v_i,e_{j,1})^{\sf{T}}\oplus\cdots\oplus \mathbf{f}_{\operatorname{el}}(v_i,e_{j,N-1})^{\sf{T}}]^{\sf{T}},\nonumber
    \end{aligned}
\end{equation}
where $e_{j,k}$ is the $k$-th query elements of $Q_j$.
We compute the BON features of node $v_i$ via max-pooling the initial embeddings of the adjacent neighboring nodes $\mathcal{N}(v_i)$:
\begin{equation}
    \begin{aligned}
    \mathbf{g}_{\operatorname{BON}}(v_i)=\max_{v\in\mathcal{N}(v_i)}\mathbf{g}(v).
    \end{aligned}
\end{equation}

\subsubsection{Graph Attention Network}\label{subsubsec:method43}
\begin{figure}[t]
\centering
\subfigure[Bag-of-neighbors.]{
\label{fig:bon}
\includegraphics[width=0.45\linewidth]{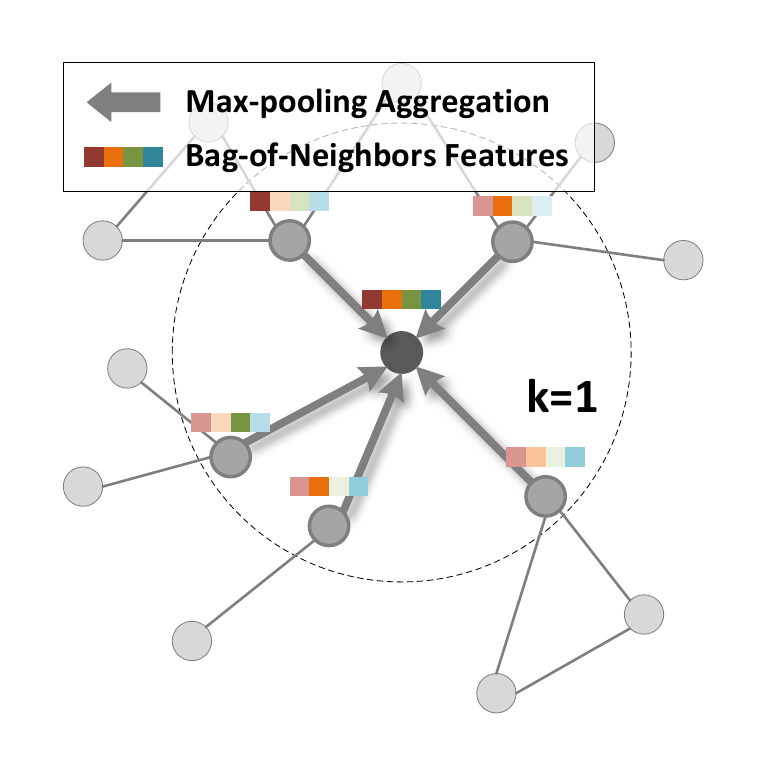}}
\subfigure[Graph attention networks.]{
\label{fig:gat}
\includegraphics[width=0.45\linewidth]{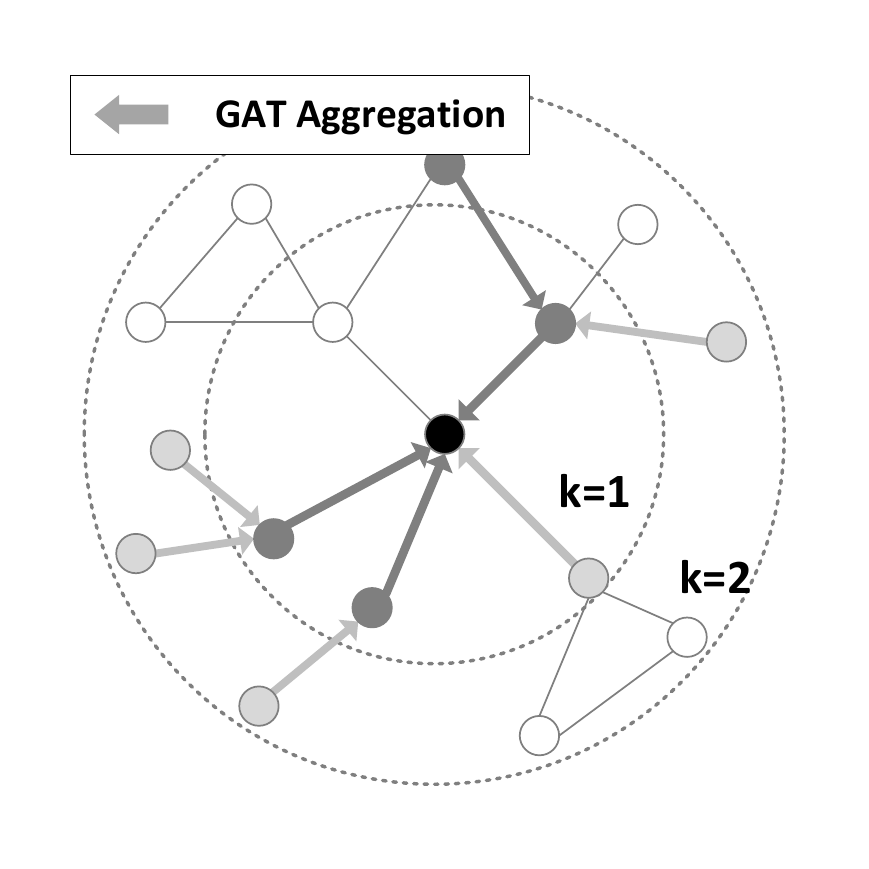}}
\vspace{-2ex}
\caption{ Illustration of (a) bag-of-neighbors features and (b) GAT.
For both sub-figures, nodes in darker grey contribute more during the aggregation.
}\label{fig:bongat}
\end{figure}

Using BON features alone may not be  expressive enough when there is query information missing in the $1$st-order neighborhood.
To include multi-hop relations from distant nodes, we apply a graph attention network~\cite{velivckovic2018graph} to aggregate such information (Figure~\ref{fig:gat}).
GAT first computes the normalized attention coefficients $\alpha_{i,j}^{(l)}$ between node $i$ in the multi-modal correlation graph and its neighboring node $j\in\mathcal{N}(i)$ in the $l$-th layer:
\begin{equation*}
    \begin{aligned}
    \alpha_{i,j}^{(l)}=\frac{\exp(\sigma(\mathbf{a}^{\sf{T}}[\mathbf{W}^{(l)}\mathbf{h}_i^{(l)}\oplus\mathbf{W}^{(l)}\mathbf{h}_j^{(l)}])}{\sum_{k\in\mathcal{N}(i)}\exp(\sigma(\mathbf{a}^{\sf{T}}[\mathbf{W}^{(l)}\mathbf{h}_i^{(l)}\oplus\mathbf{W}^{(l)}\mathbf{h}_k^{(l)}])},
    \end{aligned}
\end{equation*}
where $\mathbf{h}_i^{(l)}$ is the $l$-th layer hidden features of node $i$, $\mathbf{W}$ is a learnable weight matrix, $\mathbf{a}$ is trainable weight vector parameters, and $\sigma(\cdot)$ is the $\operatorname{LeakyReLU}(\cdot)$ activation function.
The initial node embeddings are the bag-of-neighbors feature embeddings, \ie, $\mathbf{h}_i^{(0)}=\mathbf{g}_i^{\operatorname{BON}}$.
Then, we aggregate the neighbor embeddings as the new $(l+1)$-th layer node embeddings via computing a weighted sum based on the computed attention coefficients:
\begin{equation}
    \begin{aligned}
    \mathbf{h}_i^{(l+1)}=\sigma \left(\sum_{j\in\mathcal{N}(i)}\alpha_{ij}^{(l)}\mathbf{W}^{(l)}\mathbf{h}_i^{(l)}\right).
    \end{aligned}
\end{equation}
For an $L$-layer GAT, the updated node embedding is denoted as  $\mathbf{g}_{\operatorname{BON}}'(i)=\mathbf{h}_i^{(L)}$.

\subsubsection{Multi-View Aggregation}\label{subsubsec:method44}
Although the GAT-propagated BON representations can enable the
model to extract more answers from tables, they can fall short  on paragraphs because of the lack of encoding the original semantic information in BERT embeddings.
Thus, aside from the graph-based branch introduced in \cref{subsubsec:method41}, \cref{subsubsec:method42}, and \cref{subsubsec:method43}, we add another branch based on the BERT representations of these nodes.
However, simply concatenating the BON features and the BERT embeddings might lead to several drawbacks: (1) one of the views dominating the other one during training; (2) different features with different dimensionality, making it difficult to learn a unifed classifier on the concatenation.
Thus, we design two simple classifiers and make them mutually enhance each other during the entity selection: 
(1) one classifier based on the concatenation of the entity nodes' and query
elements' BERT embeddings, and (2) the other classifier based on the GAT-updated BON features.
Given the node $v_i$ and the query $Q_j$, and using feedforward neural networks (FFNN) as classifiers, we have:
\begin{equation}
    \begin{aligned}
    \mathbf{h}_{\operatorname{s}}=[\mathbf{h}({e_{j,1}})^{\sf{T}}\oplus\cdots&\oplus\mathbf{h}({e_{j,N-1}})^{\sf{T}}\oplus\mathbf{h}(v_i)^{\sf{T}}]^{\sf{T}},\\
    \mathbf{\hat{y}}^{(\operatorname{s})}=\operatorname{FFNN}(\mathbf{h}_{\operatorname{s}}),\ &\mathbf{\hat{y}}^{(\operatorname{n})}=\operatorname{FFNN}(\mathbf{g}'_{\operatorname{BON}}(v_i)).
    \end{aligned}
\end{equation}
Then, we average the scores from simple classifiers as the prediction of the final aggregated classifier:
\begin{equation}
    \begin{aligned}
    \mathbf{\hat{y}}^{\operatorname{low}}=\operatorname{Softmax}\left(\frac{1}{2}(\mathbf{\hat{y}}^{(\operatorname{s})}+\mathbf{\hat{y}}^{(\operatorname{n})})\right).
    \end{aligned}\label{eq:lowagg}
\end{equation}

%% file: 33trainobj.tex
\subsection{Training Objective}
Given a document $D_k$ and a query $\mathcal{Q}_j$, $\mathbf{y}_{jk}^{\operatorname{high}}$ and $\mathbf{y}_{jk}^{\operatorname{low}}$ indicate the ground-truth label of the correct component and entity for high-level component retrieval and low-level entity extraction, while $\mathbf{\hat{y}}_{jk}^{\operatorname{high}}$ and $\mathbf{\hat{y}}_{jk}^{\operatorname{low}}$ indicate the predictions from the component retriever and entity extractor. We define the following training objectives:

\paragraph{High-Level Component Retrieval}
We use the traditional classification loss, $\ell_{\operatorname{CE}}(\mathbf{y},\mathbf{\hat{y}})=\sum_i-y_i\log\hat{y}_i-(1-y_i)\log(1-\hat{y}_i)$, as the high-level model training objective:
\begin{equation}\label{eq:lossh}
    \begin{aligned}
    \ell_{\operatorname{high}}&=\sum_{k=1}^{M}\sum_{j=1}^{|\mathcal{Q}|}\ell_{\operatorname{CE}}(\mathbf{y}_{jk}^{\operatorname{high}},\mathbf{\hat{y}}_{jk}^{\operatorname{high}}). 
    \end{aligned}
\end{equation}

\paragraph{Low-Level Entity Selection}
The training objective
for the low-level entity classifiers (\cref{subsubsec:method44})  is separated
into three parts: 
(1) the aggregated classification loss for  the aggregated model: 
\begin{equation}\label{eq:loss1}
    \begin{aligned}
    \ell_1&=\sum_{k=1}^M\sum_{j=1}^{|\mathcal{Q}|}\ell_{\operatorname{CE}}(\mathbf{y}_{jk}^{\operatorname{low}},\mathbf{\hat{y}}_{jk}^{\operatorname{low}}),
    \end{aligned}
\end{equation}
(2) the classification loss for the two subclassifiers:
\begin{equation}\label{eq:loss2}
    \begin{aligned}
    \ell_2&=\sum_{k=1}^M\sum_{\operatorname{v}\in\{\operatorname{s},\operatorname{n}\}}\sum_{j=1}^{|\mathcal{Q}|}\ell_{\operatorname{CE}}(\mathbf{y}_{jk}^{\operatorname{low}},\mathbf{\hat{y}}_{jk}^{(\operatorname{v})}),
    \end{aligned}
\end{equation}
(3) the consistency loss between the two subclassifiers to encourage them to reach a consensus:
\begin{equation}\label{eq:loss3}
    \begin{aligned}
    \ell_3&=\sum_{k=1}^M\sum_{\operatorname{u},\operatorname{v}\in\{\operatorname{s},\operatorname{n}\}}\sum_{j=1}^{|\mathcal{Q}|}\|\mathbf{\hat{y}}_{jk}^{(\operatorname{u})}-\mathbf{\hat{y}}_{jk}^{(\operatorname{v})}\|_2^2. 
    \end{aligned}
\end{equation}

The overall objective of the low-level entity extractor is then:
\begin{equation}\label{eq:lossfinal}
    \begin{aligned}
    \ell_{\operatorname{low}}=\ell_1+\lambda\ell_2+\mu\ell_3, 
    \end{aligned}
\end{equation}
where $\lambda$ and $\mu$ are pre-defined 
balancing 
hyper-parameters. 

%% file: 4experiments.tex
\section{Experiments}\label{sec:exp}
\input{41setup}\label{subsec:exp1}
\input{42results}\label{subsec:exp2}

%% file: 41setup.tex
\subsection{Experimental Setup}
\paragraph{Datasets.}
We evaluate our model on three SciIE datasets: 
(1) \textbf{SciREX}~\cite{jain2020scirex}  contains 438 annotated full-length machine learning papers;
(2) \textbf{PubMed}~\cite{jia2019document} 
contains 5688 annotated full-length biochemical papers;
(3) \textbf{NLP-TDMS (Full)}~\cite{hou2019identification} contains 332 unannotated full-length natural language processing papers (see Appendix~\ref{appendix:dataset} for more details). 
We extend the original datasets to include both text and tables from the LaTeX or PDF files.
Our experiments show that the domain-specific BERTs work better than the general domain BERT model.
For SciREX and TDMS-NLP, we use SciBERT~\cite{beltagy-etal-2019-scibert} as the encoder for all methods; for PubMed, we use ClinicalBERT~\cite{alsentzer-etal-2019-publicly}. 

\paragraph{Component Retrieval Baselines.}
Our baselines contain
(1) \textit{sparse retrieval methods}: \textbf{TF-IDF}~\cite{aizawa2003information}, \textbf{BM25}~\cite{robertson2009probabilistic};
(2) \textit{entity-based methods}: \textbf{Entity Cosine Similarities (ECS)}, \textbf{Deep Entity Cosine Similarities (DECS)};
(3) \textit{embedding-based methods}: \textbf{BERT-Matching (BERT-M)}, \textbf{BERT-Entailment (BERT-E)}~\cite{nogueira2019passage, nie2019revealing}, \textbf{Recurrent Retriever (RR)}~\cite{asai2019learning}, \textbf{Dense Passage Retrieval (DPR)}~\cite{karpukhin2020dense}.

\paragraph{Entity Selection Baselines.}
For the low-level model, we restrict the searching space to the ground-truth paragraph/table that contains the answer.
The baselines include
(1) \textit{BERT-based methods}: \textbf{BERT-Base}, \textbf{SciREX}~\cite{jain2020scirex};
(2) \textit{graph-based methods}: \textbf{Graph Convolutional Network (GCN)}~\cite{kipf2017semi}, \textbf{GAT}~\cite{velivckovic2018graph}, \textbf{Heterogeneous Document-Entity (HDE) graph}~\cite{tu2019multi};
(3) \textit{pre-trained language models}: \textbf{TAPAS}~\cite{herzig2020tapas}, \textbf{TDMS-IE}~\cite{hou2019identification}.

\paragraph{Overall Baselines.}
(1) \textbf{BERT-Base} model searching in the whole document; (2) \textbf{GCN} and (3) \textbf{GAT} testing the performance of our proposed graph in the whole document; (4) \textbf{BERT-Entailment+Base} combining the best baselines for both high- and low-level stage (see Appendix \ref{appendix:baseline} for more details about baselines).


\paragraph{Metrics.}
Following existing works~\cite{karpukhin2020dense}, we use (1) Accuracy (Acc), (2) Mean Reciprocal Rank (MRR), and (3) Top-k Hit Rate (Hit@K) with $k=2,3,5$ for evaluating both high- and low-level models (see Appendix \ref{appendix:eval}).

%% file: 42results.tex
\input{421mainresults}

\input{422ablation}

\input{423parameter}

\input{424case}

%% file: 421mainresults.tex
\subsection{Comparison with Baselines}
Table~\ref{tab:main1}--\ref{tab:main3} present the average performances over multiple random trials~\footnote{The standard deviation is reported in Appendix~\ref{appendix:std}.}. 
\ours consistently
outperforms the strongest baselines by $9.01\%$, $6.81\%$, $10.25\%$ in  Acc and 
$4.38\%$, $12.01\%$, $8.27\%$ in MRR on the three datasets at all levels. 
For the other ranking metrics of hit rate, \ours also show marginal improvements compared with baselines.

As \ours-H employs both component-level semantic features and entity-level matching features, its high-level performance exceeds that of the sparse and dense retrieval baselines, which captures only single-sided information.
For low-level, the embedding-based methods and pretrained-LMs only take advantage of the latent semantic information, while the graph-based methods only focus on the graph topology and can easily suffer from  over-smoothing.
Compared with these baselines, \ours-L shows better performances with the multi-view aggregation of GAT-propagated BON features and BERT embeddings. 

Comparing the performance gains for the low-level extraction, we find that the components of \ours-L contribute variously on different datasets. 
In SciREX, with most of the questioned scores hidden in the tables, the GAT-propagated BON features work better in discriminating the table cells and numeric values. 
For PubMed, the targets mostly appear in text rather than tables.
Thus, the semantic information in BERT embeddings contributes more to the performance increase. 
NLP-TDMS is a benchmark dataset that includes multiple relevant choices for a given query, the ambiguity of which hurts the performance of all models.

\input{40maintables}

%% file: 40maintables.tex
\begin{table*}[htb]
\centering
\footnotesize
\setlength\tabcolsep{2pt}
\begin{tabular}{@{}lccccccccccccccc@{}}
\toprule
                  & \multicolumn{5}{c|}{\textbf{SciREX}}                                                                         & \multicolumn{5}{c|}{\textbf{PubMed}}    &     \multicolumn{5}{c}{\textbf{NLP-TDMS}}                                   \\ \midrule
\textbf{Methods}                   & \textbf{Acc}    & \textbf{MRR}    & \textbf{Hit@2}  & \textbf{Hit@3}  & \multicolumn{1}{c|}{\textbf{Hit@5}}  & \textbf{Acc}    & \textbf{MRR}    & \textbf{Hit@2}  & \textbf{Hit@3}  & \multicolumn{1}{c|}{\textbf{Hit@5}} & \textbf{Acc} & \textbf{MRR} & \textbf{Hit@2} & \textbf{Hit@3} & \textbf{Hit@5} \\ \midrule
TF-IDF & 9.31 & 17.27 & 12.64 & 14.48 & \multicolumn{1}{c|}{17.28} & 30.41 & 46.60 & 45.43 & 54.19 & \multicolumn{1}{c|}{65.71} & 9.71 & 18.31 & 13.31 & 17.27 & 25.76 \\
BM25 & 27.44 & 42.86 & 39.90 & 50.75 & \multicolumn{1}{c|}{63.81} & 28.29 & 44.04 & 42.43 & 50.56 & \multicolumn{1}{c|}{62.08} & 13.38 & 24.19 & 19.95 & 24.51 & 33.14 \\ \midrule
ECS	& 16.37 & 29.11 & 20.34 & 25.03 & \multicolumn{1}{c|}{42.00} & 17.65 & 32.50 & 27.91 & 34.92 & \multicolumn{1}{c|}{47.93} & 22.84 & 36.00 & 31.28 & 38.88 & \multicolumn{1}{c}{45.52} \\
DECS & 25.82 & 43.48 & 37.47 & 52.68 & \multicolumn{1}{c|}{71.89} & 37.36 & 45.57 & 43.93 & 52.38 & \multicolumn{1}{c|}{64.14} & 28.20 & 46.87 & 45.67 & 59.56 & 68.52 \\ \midrule
BERT-M & 52.38 & 67.54 & 53.97 & 61.11 & \multicolumn{1}{c|}{76.19} & 45.78 & 61.13 & 62.58 & 70.84 & \multicolumn{1}{c|}{80.78} & 46.18 & 62.39 & 64.83 & 73.72 & 83.98 \\ 
BERT-E & 60.59 & 75.98 & 82.04 & 88.92 & \multicolumn{1}{c|}{98.37} & 47.35 & 63.28 & \textbf{66.29} & 73.30 & \multicolumn{1}{c|}{81.64} & 50.77 & 66.97 & 64.62 & 86.15 & 86.15 \\
RR & 25.42 & - & - & - & \multicolumn{1}{c|}{-} & 35.29 & - & - & - & \multicolumn{1}{c|}{-} & 31.87 & - & - & - & - \\
DPR & 53.47 & 50.26 & 58.42 & 74.25 & \multicolumn{1}{c|}{88.96} & 45.31 & 61.47 & 64.46 & 73.22 & \multicolumn{1}{c|}{81.48} & 57.14 & 72.98 & 76.19 & 85.71 & 97.62 \\ \midrule
\textbf{ReSel-H} & \textbf{71.48} & \textbf{82.48} & \textbf{85.59} & \textbf{93.01} & \multicolumn{1}{c|}{\textbf{98.61}} & \textbf{49.02} & \textbf{63.67} & 66.21 & \textbf{74.43} & \multicolumn{1}{c|}{\textbf{83.81}} & \textbf{71.62} & \textbf{79.21} & \textbf{83.33} & \textbf{91.87} & \textbf{99.36} \\ \bottomrule
\end{tabular}
\caption{The performance of different methods for retrieving high-level components. We
measure the performance of different methods in retrieving the ground-truth
components in terms of accuracy, MRR, and top-$k$ hit ratios.
}\label{tab:main1}
\end{table*}

\begin{table*}[htb]
\centering
\footnotesize
\setlength\tabcolsep{2pt}
\begin{tabular}{@{}lccccccccccccccc@{}}
\toprule
                  & \multicolumn{5}{c|}{\textbf{SciREX}}                                                                         & \multicolumn{5}{c|}{\textbf{PubMed}}  & \multicolumn{5}{c}{\textbf{NLP-TDMS}}                                          \\ \midrule
\textbf{Methods}                   & \textbf{Acc}    & \textbf{MRR}    & \textbf{Hit@2}  & \textbf{Hit@3}  & \multicolumn{1}{c|}{\textbf{Hit@5}} & \textbf{Acc}    & \textbf{MRR}    & \textbf{Hit@2}  & \textbf{Hit@3}  & \multicolumn{1}{c|}{\textbf{Hit@5}} & \textbf{Acc} & \textbf{MRR} & \textbf{Hit@2} & \textbf{Hit@3} & \textbf{Hit@5} \\ \midrule
Base & 14.72 & 25.81 & 20.96 & 28.61 & \multicolumn{1}{c|}{40.60} & 72.50 & 77.36 & 82.01 & 89.44 & \multicolumn{1}{c|}{95.87} & 9.37 & 21.42 & 15.62 & 21.88 & 33.54 \\
SciREX & 14.21 & 23.25 & 20.42 & 26.56 & \multicolumn{1}{c|}{37.23} & 52.44 & 63.42 & 70.97 & 80.36 & \multicolumn{1}{c|}{91.45} & 11.35 & 23.75 & 17.86 & 26.33 & 41.09 \\ \midrule
GCN & 10.74 & 21.60 & 16.72 & 22.87 & \multicolumn{1}{c|}{31.00} & 57.36 & 72.37 & 77.09 & 87.02 & \multicolumn{1}{c|}{94.61} & 12.79 & 21.60 & 17.55 & 20.31 & 37.24 \\
GAT & 12.09 & 21.72 & 17.39 & 22.28 & \multicolumn{1}{c|}{34.23} & 57.44 & 71.35 & 77.13 & 86.93 & \multicolumn{1}{c|}{95.95} & 14.69 & 25.12 & 17.62 & 33.61 & 42.58 \\
HDE	& 14.78 & 24.79 & 21.12 & 27.07 & \multicolumn{1}{c|}{33.70} & 60.73 & 72.90 & 81.32 & 87.77 & \multicolumn{1}{c|}{93.93} & 15.38 & 28.74 & 19.23 & 32.47 & 46.75 \\ \midrule
TAPAS & 25.45 & - & - & - & \multicolumn{1}{c|}{-} & 8.63 & - & - & - & \multicolumn{1}{c|}{-} & 23.79 & - & - & - & - \\
TDMS-IE & 18.41 & - & - & - & \multicolumn{1}{c|}{-} & 6.42 & - & - & - & \multicolumn{1}{c|}{-} & 13.44 & - & - & - & - \\ \midrule
\textbf{ReSel-L} & \textbf{41.68} & \textbf{51.45} & \textbf{49.30} & \textbf{55.70} & \multicolumn{1}{c|}{\textbf{65.50}} & \textbf{74.71} & \textbf{79.57} & \textbf{84.22} & \textbf{91.06} & \multicolumn{1}{c|}{\textbf{97.57}} & \textbf{25.77} & \textbf{36.91} & \textbf{25.91} & \textbf{40.64} & \textbf{49.72}\\ \bottomrule
\end{tabular}
\caption{The performance of different low-level methods in extracting the target entities. 
}\label{tab:main2}
\end{table*}

\begin{table*}[htb]\small
\centering
\vspace{-2ex}
\setlength\tabcolsep{2pt}
\begin{tabular}{@{}lccccccccccccccc@{}}
\toprule
                  & \multicolumn{5}{c|}{\textbf{SciREX}}                                                                         & \multicolumn{5}{c|}{\textbf{PubMed}}  & \multicolumn{5}{c}{\textbf{NLP-TDMS}}                                          \\ \midrule
\textbf{Methods}                   & \textbf{Acc}    & \textbf{MRR}    & \textbf{Hit@2}  & \textbf{Hit@3}  & \multicolumn{1}{c|}{\textbf{Hit@5}} & \textbf{Acc}    & \textbf{MRR}    & \textbf{Hit@2}  & \textbf{Hit@3}  & \multicolumn{1}{c|}{\textbf{Hit@5}} & \textbf{Acc} & \textbf{MRR} & \textbf{Hit@2} & \textbf{Hit@3} & \textbf{Hit@5} \\ \midrule
Base & 6.53 & 11.35 & 9.42 & 10.42 & \multicolumn{1}{c|}{15.15} & 26.63 & 30.16 & 26.06 & 33.56 & \multicolumn{1}{c|}{43.21} & 3.13 & 5.62 & 4.14 & 4.80 & 6.66 \\
GCN & 4.03 & 5.87 & 5.54 & 6.11 & \multicolumn{1}{c|}{10.11} & 16.63 & 25.95 & 21.42 & 28.47 & \multicolumn{1}{c|}{40.41} & 7.61 & 16.73 & 9.18 & 16.83 & 20.92 \\
GAT & 8.44 & 11.93 & 9.73 & 10.19 & \multicolumn{1}{c|}{13.47} & 16.80 & 26.21 & 22.79 & 29.60 & \multicolumn{1}{c|}{38.59} & 9.82 & 16.24 & 13.13 & 14.42 & 15.79 \\
BERT-E+B & 16.27 & 22.85 & 18.02 & 23.98 & \multicolumn{1}{c|}{31.09} & 29.30 & 37.59 & 41.17 & 43.36 & \multicolumn{1}{c|}{\textbf{53.50}} & 7.58 & 11.86 & 7.18 & 11.48 & 13.34 \\
\textbf{ReSel} & \textbf{38.69} & \textbf{43.66} & \textbf{42.90} & \textbf{46.35} & \multicolumn{1}{c|}{\textbf{48.51}} & \textbf{33.73} & \textbf{40.77} & \textbf{42.38} & \textbf{44.99} & \multicolumn{1}{c|}{46.19} & \textbf{13.71} & \textbf{17.55} & \textbf{15.06} & \textbf{18.80} & \textbf{22.87} \\ \bottomrule
\end{tabular}
\caption{The overall document-level extraction performance of different methods. 
}\label{tab:main3}
\end{table*}

%% file: 422ablation.tex
\subsection{Ablation Studies}


We conduct ablation studies on SciREX and present the results in Table~\ref{tab:ablation}.

\begin{table}[h]
\centering
\footnotesize
\vspace{-2ex}
\setlength\tabcolsep{2pt}
\begin{tabular}{@{}l|ccccc@{}}
\toprule
\textbf{High-Level Models}               & \textbf{Acc}               & \textbf{MRR}               & \textbf{Hit@2}             & \textbf{Hit@3}             & \textbf{Hit@5}             \\ \midrule
ReSel-H w/o CS &   34.12              &   55.02                 &  57.93                     & 73.01                     &    81.74                  \\
ReSel-H w/o ES &        58.73                    &    72.03                    &   72.22                   &  79.36                    &             92.06         \\
ReSel-H w/o EL  &  52.38                    &   67.88                  &   66.66                         &    76.98                        &       \textbf{96.03 }                    \\
\textbf{ReSel-H}                            & \textbf{64.28}  &  \textbf{74.90}  &  \textbf{74.60}  &    \textbf{80.15}& 92.06   \\ \midrule
\textbf{Low-Level Models}                & \textbf{Acc}               & \textbf{MRR}               & \textbf{Hit@2}             & \textbf{Hit@3}             & \textbf{Hit@5}             \\ \midrule
ReSel-L w/o BON                    & 15.86                     & 26.15                     & 21.37                     & 25.51                     & 37.93                     \\
ReSel-L w/o GAT                             & 41.37                     & 45.50                     & 41.37                     & 41.37                     & 57.24                     \\
ReSel-L w/o OS       & 48.28 & 53.08 & 51.03 & 55.86 & 60.69 \\
ReSel-L w/o MVA                            &   19.31                 & 27.65                    & 24.13                        &    27.58                        &  33.10                          \\
\textbf{ReSel-L}                            & \textbf{51.72}            & \textbf{57.60}            & \textbf{56.55}            & \textbf{60.00}            & \textbf{64.83}            \\ \bottomrule
\end{tabular}
\caption{Performance comparison of ablation study.}\label{tab:ablation}
\end{table}

\paragraph{Multi-View Features.}
\textbf{CS features} enable \ours-H to measure the matching scores between the main topics of the components and the query.
Removing this feature will cause larger performance degradation than removing the other two features, indicating that for high-level retrieval, component-level information guides primary retrieval, while entity-level information refines it.
At the entity level, \textbf{ES features} and \textbf{EL features} allow the model to capture semantic and lexical relevance between paragraph/table and query elements.
By removing these entity-level features, the model will rely solely on BERT embeddings, which are less expressive for lengthy paragraphs and tables.

\paragraph{Graph-based Branch.}
When we replace the \textbf{BON features} with BERT embeddings, \ours-L's performance drops a lot. 
This demonstrates that BERT embeddings cannot discriminate table entities well, especially for numeric values and adjacent table cells.
In contrast, the BON features, encoding neighboring information and graph topology, can distinguish such entities.
On the other hand, \textbf{GAT} also improves the performance. 
As BON features are based on adjacent nodes, the pooling aggregation is only over 1st-order neighbors. Thus, GAT can complement BON features by propagating distant neighborhood information.


\paragraph{Original Semantic Branch.}
For low-level model, including BERT-based \textbf{Original Semantic (OS) Features} improves the performance. 
Although the improvement is marginal on SciREX, it is notable on other datasets (e.g., PubMed) where many answers reside in paragraphs.
Removing \textbf{Multi-View Aggregation (MVA)}
 will make \ours-L's performance decrease significantly. 
This is because  when simply concatenating the BERT embeddings and the GAT-propagated BON features, the BERT embeddings (which has much higher dimensionality) can 
dominate the learning process.

%% file: 423parameter.tex
\subsection{Parameter Studies}
\begin{figure}[h]
    \centering
    \includegraphics[width=\linewidth]{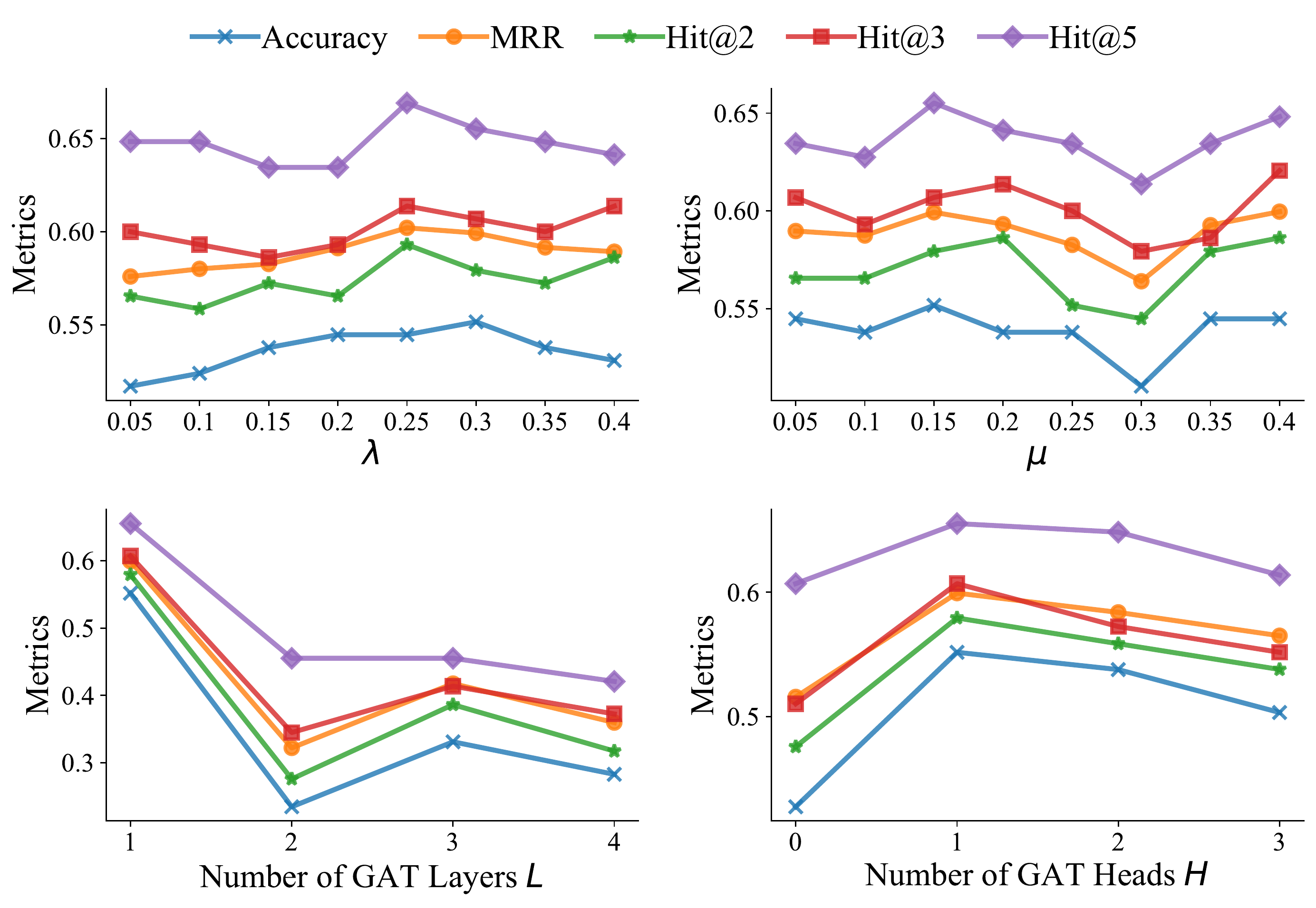}
    \caption{Parameter study of \ours on SciREX.
    }
    \label{fig:para}
\end{figure}

Figure~\ref{fig:para} shows our parameter study results.



\paragraph{$\lambda$ and $\mu$.}
The loss $\ell_1$ in the aggregated classifier in Eq.~\eqref{eq:lowagg} plays the leading role in training objective.
When $\lambda$ is too small or $\mu$ is too large, the regularization of consistency between two classifiers will contribute more than their respective classification loss, making them more intended to generate incorrect-but-same predictions;
Conversely, when $\lambda$ is too large or $\mu$ is too small, the classifiers begin to generate biased predictions, making the aggregation deteriorate to a mere average.

\paragraph{$L$ and $H$.}
The number of GAT layers $L$ determines the depth of neighboring information on the graph, also known as the orders of neighbors in aggregation. 
When $L$ is increasing, we will aggregate more common neighbors for adjacent nodes, making it easier for GAT to fall into over-smoothing.
The width of neighboring information on the graph is dictated by the amount of relationships we encode from neighbors.
When we increase the number of attention heads $H$, GAT will learn and combine several sets of attention scores on the neighboring nodes, which can also include more irrelevant or misleading information from them. 
Besides, whichever $H$ or $L$ is increasing, the model needs to train more parameters, taking more time and data.

%% file: 424case.tex

\subsection{Case Study}
\begin{figure}[h]
\centering
\includegraphics[width=\linewidth]{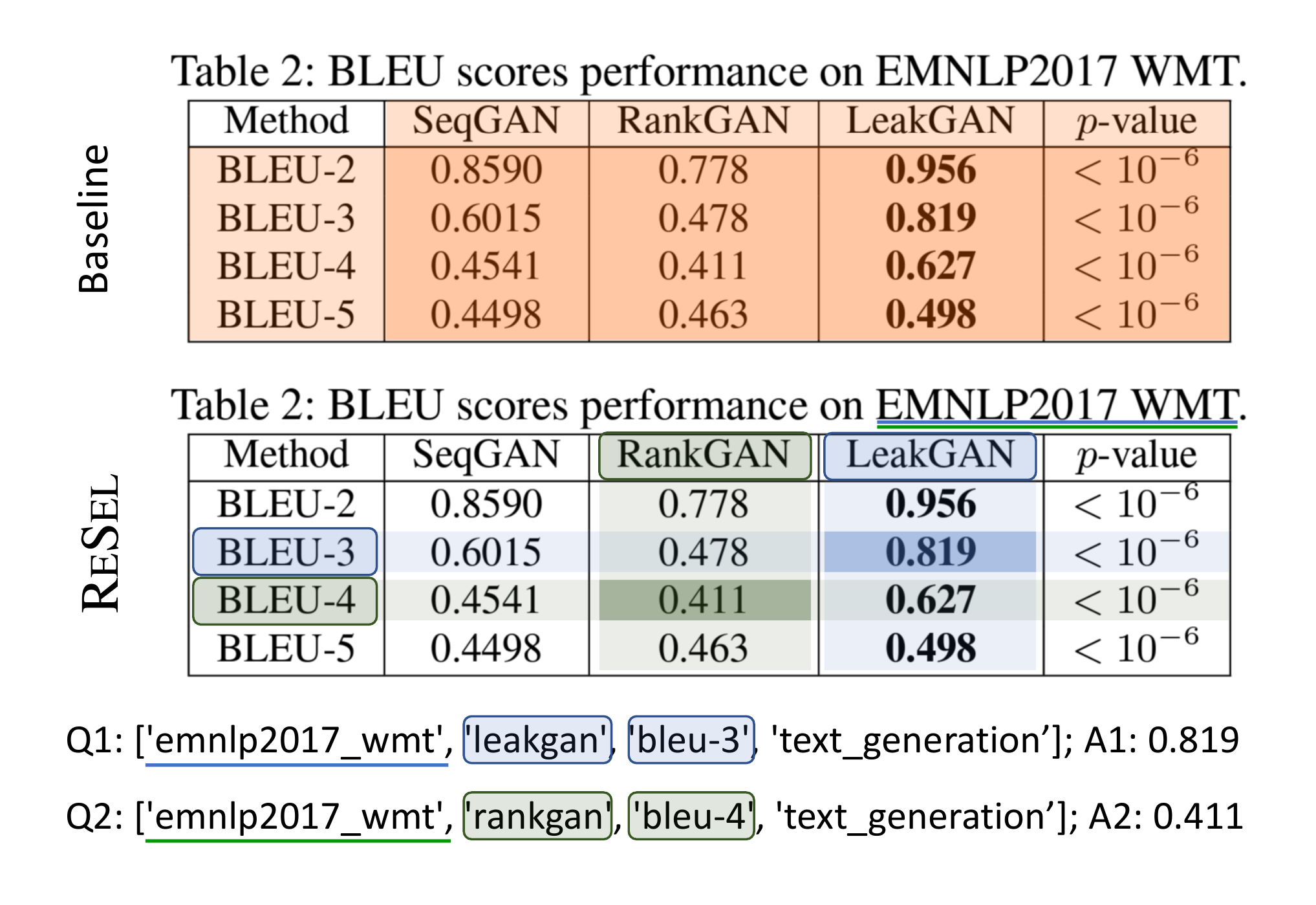}
\vspace{-2.5ex}
\caption{The case study of predictions made by the baseline GCN model and \ours model.
}
\vspace{-2ex}
\label{fig:case}
\end{figure}

Figure~\ref{fig:case} shows a representative example to illustrate the efficacy of \ours.
It shows the predictions from GCN baseline and \ours for two queries on the same document.
The darker the color is on the table cell, the higher prediction score we obtain for it.
We can clearly see that BERT embeddings alone cannot distinguish which numerical value is the final answer.
The graph-propagated embeddings are brought even closer in semantic space by linking with related items in column and row headers. 
However, with the BON features and graph topology, \ours can distinguish different values in the table and make correct selections.


%% file: 5conclusion.tex
\section{Conclusion}

We proposed \ours, a two-stage method for 
 $N$-ary relation extraction jointly from scientific text and tables.
 \ours consists of two key components: a high-level component retriever and a low-level entity extractor.
The multiple features defined in the high-level retriever enables our model to leverage semantic and lexical information from both paragraphs/tables and entities.
For low-level entity extractor, the multi-view aggregation effectively encodes both the topology information from the graph and the semantic information from pre-trained BERT embeddings. 
Extensive experiments on three datasets show that \ours consistently outperforms all baseline models significantly. 


%% file: 6limit.tex
\section*{Limitations}

While \ours has demonstrated superior performance compared with the state-of-the-art baselines, it has several limitations that can be addressed in the future.
First, although \ours extends the capability of previous $N$-ary relation extraction to both text and tables, it cannot extract from images---another important data modality in scientific articles.
This necessitates augment \ours with optical character recognition (OCR) techniques to parse images and jointly extract from the text, table, and image modalities.
Second, we found the datasets for SciIE are limited and expensive to curate, especially as we aim to expand to include images.
Accurate annotation for multi-modal SciIE is time-consuming and needs  more future collaborative efforts from related  communities.
Third, currently \ours has not modeled the layout information (\eg, font style, font size, \etc), which may also contain some clues for intra- and inter-modality relations.
Some existing studies~\cite{xu2020layoutlm,xu-etal-2021-layoutlmv2,xu2021layoutxlm} have worked on pre-training models that encode the layout information, which can be interesting to be combined with \ours.


%% file: ack.tex
\section*{Acknowledgements}
This work was supported in part by NSF (IIS-2008334, IIS-2106961, CAREER IIS-2144338), ONR (MURI N00014-17-1-2656), and Kolon Industries.

%% file: appendices/appendix.tex
\begin{table*}[htb]
\footnotesize
\centering
\begin{tabular}{@{}lccc@{}}
\toprule
\textbf{Dataset}           & \textbf{SciREX}                                                 & \textbf{PubMed }                             & \textbf{NLP-TDSM}                                       \\ \midrule
Query Elements    & <Task, Method, Dataset, Metric> & <Drug, Gene> & <Task, Dataset, Metric> \\
Answer Type       & Score                                                  & Mutation                            & Score                                          \\
Training/Val/Test & 263/88/87                                              & 2366/592/799                        & 200/66/66                                      \\ \bottomrule
\end{tabular}
\caption{Datasets details. The numbers in training/val/test set split are the numbers of full-length scientific paper documents.}
\label{tab:data}
\end{table*}

\section{Methodology Computational Details}
\subsection{Query and Component Encoder}
\label{appendix:ce}
\noindent$\bullet~$\textbf{Query Encoder} 
Each query $Q_j$ contains $(N-1)$ elements $[e_{j,1},\cdots,$ $e_{j,N-1}]$.
To generate a more understandable natural language sequence for the BERT encoder, we re-formulate the query into a question $[q_{j,1},\cdots,q_{j,M_j}]$, where $M_j$ is the number of words in the generated question.
In this way, we are able to use the [CLS] token embedding as the query embedding:

\begin{small}
\begin{equation}
    \begin{aligned}
    \mathbf{h}(Q_j)=\operatorname{BERT_{[CLS]}}([q_{j,1},\cdots,q_{j,M_j}]).
    \end{aligned}
\end{equation}
\end{small}

\begin{small}
\begin{equation}
    \begin{aligned}
    \{\mathbf{h}(q_{j,1}),\cdots,\mathbf{h}({q_{j,M_j}})\}&=\operatorname{BERT}([q_{j,1},\cdots,q_{j,M_j}]).
    \end{aligned}
\end{equation}
\end{small}

By averaging the embeddings of words that are related to the query elements, we obtain the $a$-th query element embeddings $\mathbf{h}({e_{j,a}}),e_{j,a}\in Q_j$:
\begin{equation}
    \begin{aligned}
    \mathbf{h}(e_{j,a})&=\frac{\sum_{k=1}^{M_j}\mathbf{h}({q_{j,k}})\cdot\mathbb{I}(q_{j,k}\in e_{j,a})}{\sum_{k=1}^{M_j}\mathbb{I}(q_{j,k}\in e_{j,a})}.
    \end{aligned}
\end{equation}

\noindent$\bullet~$\textbf{Component Encoder} 
As is mentioned in \cref{sec:pf}, each component in a document can be denoted as a sequence of words $[w_{i,1},\cdots,w_{i,N_i}]$.
Then, we directly encode the paragraph embedding $\mathbf{h}({C_i})$, the included
word embeddings $\{\mathbf{h}({w_{i,1}}),\cdots,\mathbf{h}({w_{i,N_i}})\}$, and the
averaged entity embeddings $\mathbf{h}({m_{i,b}})$, where $m_{i,b}\in C_i$
indicates the $b$-th entity extracted from the component $C_i$:

\begin{small}
\begin{equation}
    \begin{aligned}
    \mathbf{h}({C_i})&=\operatorname{BERT_{[CLS]}}([w_{i,1},\cdots,w_{i,N_i}]),
    \end{aligned}
\end{equation}
\begin{equation}
    \begin{aligned}
    \{\mathbf{h}({w_{i,1}}),\cdots,\mathbf{h}({w_{i,N_i}})\}&=\operatorname{BERT}([w_{i,1},\cdots,w_{i,N_i}]),
    \end{aligned}
\end{equation}
\begin{equation}
    \begin{aligned}
    \mathbf{h}({m_{i,b}})&=\frac{\sum_{k=1}^{N_i}\mathbf{h}({w_{i,k}})\cdot\mathbb{I}(w_{i,k}\in m_{i,b})}{\sum_{k=1}^{N_i}\mathbb{I}(w_{i,k}\in m_{i,b})}.
    \end{aligned}
\end{equation}
\end{small}

%% file: appendices/appendix2.tex
\subsection{Text Similarities}
\label{appendix:textsim}
\noindent$\bullet~$\textbf{Levenshtein Similarity:} The string similarity based on Levenshtein Distance~\cite{levenshtein1966binary}:

\begin{small}
\begin{equation}
    \begin{aligned}
    \operatorname{Leven\_Sim}(m_{i,b},e_{j,a})=1-\frac{\operatorname{Leven\_Dist}(m_{i,b},e_{j,a})}{\max(|m_{i,b}|,|e_{j,a}|)},
    \end{aligned}
\end{equation}
\end{small}

where $\operatorname{Leven\_Dist(\cdot,\cdot)}$ refers to the Levenshtein Distance, which measures how different two strings are by counting the number of deletions, insertion, or substitutions required to transform one string to another.

\noindent$\bullet~$\textbf{Longest Common Substring:} The ratio between the length of longest common substring and the minimum length of the two strings:
\begin{equation}
    \begin{aligned}
    \operatorname{LCStr\_Sim}(m_{i,b},e_{j,a})=\frac{|\operatorname{LCStr}(m_{i,b},e_{j,a}|)}{\min(|m_{i,b}|,|e_{j,a}|)},
    \end{aligned}
\end{equation}
where $\operatorname{LCStr}(\cdot,\cdot)$ indicates the longest common substring between two given strings.

\noindent$\bullet~$\textbf{Longest Common Subsequence:} The longest common subsequence (LCS) is the longest subsequence that is common to all given strings. Different from the longest common substring, the elements of the subsequence are not needed to occupy consecutive locations within the original sequences. The ratio between the length of longest common subsequence and the minimum length of the two strings:
\begin{equation}
    \begin{aligned}
    \operatorname{LCSeq\_Sim}(m_{i,b},e_{j,a})=\frac{|\operatorname{LCSeq}(m_{i,b},e_{j,a}|)}{\min(|m_{i,b}|,|e_{j,a}|)},
    \end{aligned}
\end{equation}
where $\operatorname{LCSeq}(\cdot,\cdot)$ indicates the longest common subsequence between two given strings. 

\subsection{Cross-Modal Graph Construction}\label{appendix:graph}
\noindent$\bullet~$\textbf{Co-occurence Edge:} 
When two entity nodes $v_i$ and $v_j$ occur in the same sentence, we connect them with a co-occurence edge $E_{(v_i,v_j)}$ with weight $w_s$. If $v_i$ and $v_j$ do not co-occur in the same sentence but in two adjacent sentences, we still connect them but assign a smaller weight $w_t$ ($w_t<w_s$) to edge $E_{(v_i,v_j)}$. In practice, we set $w_t = w_s / 2$.

\noindent$\bullet~$\textbf{Co-Reference Edge} extracts the intra-paragraph relation. When two entity nodes $v_i$ and $v_j$ are referring to the same concept, we connect them with co-reference edge $E_{(v_i,v_j)}$, \eg, abbreviations and full names, common names and scientific names, \etc

\noindent$\bullet~$\textbf{Reference Edge} extracts the inter-modality relationship between paragraphs and tables. When an entity node $v_i$ occurs in a sentence with a reference mark, (\eg, ``Table. 3'', \etc), we link it to any node $v_j$ in the referenced table with a reference edge $E_{(v_i,v_j)}$.

\noindent$\bullet~$\textbf{Table-Structure Edge} extracts the intra-table relation. We connect a table-structure edge $E_{(v_i,v_j)}$ between a table cell node $v_i$ and another node $v_j$ appearing in the corresponding column header, row header, or the table caption. 

\noindent$\bullet~$\textbf{Table-Paragraph Connection} bridges the paragraph-table relation. Given an entity node $v_i$ in a paragraph and a cell node $v_j$ in a table, we place a table-paragraph connection edge $E_{(v_i,v_j)}$ between them. The weight of $E_{(v_i,v_j)}\in[0,1]$ is computed based on text similarities between the surface strings of two nodes.

%% file: appendices/appendixT.tex
\begin{table*}[htb]
\centering
\footnotesize
\vspace{-2ex}
\setlength\tabcolsep{2pt}
\begin{tabular}{@{}lccccccccccccccc@{}}
\toprule
                  & \multicolumn{5}{c|}{\textbf{SciREX}}                                                                         & \multicolumn{5}{c|}{\textbf{PubMed}}    &     \multicolumn{5}{c}{\textbf{NLP-TDMS}}                                   \\ \midrule
\textbf{Methods}                   & \textbf{Acc}    & \textbf{MRR}    & \textbf{Hit@2}  & \textbf{Hit@3}  & \multicolumn{1}{c|}{\textbf{Hit@5}}  & \textbf{Acc}    & \textbf{MRR}    & \textbf{Hit@2}  & \textbf{Hit@3}  & \multicolumn{1}{c|}{\textbf{Hit@5}} & \textbf{Acc} & \textbf{MRR} & \textbf{Hit@2} & \textbf{Hit@3} & \textbf{Hit@5} \\ \midrule
TF-IDF & 4.28 & 5.08 & 5.60 & 6.08 & \multicolumn{1}{c|}{ 6.48} & - & - & - & - & \multicolumn{1}{c|}{-} & 3.65 & 1.85 & 3.15 & 1.03 & 5.47 \\
BM25 & 10.01 & 9.30 & 11.23 & 10.14 & \multicolumn{1}{c|}{11.39} & - & - & - & - & \multicolumn{1}{c|}{-} & 2.40 & 1.45 & 1.76 & 3.46 & 3.88 \\\midrule
ECS & \textbf{2.67} & 6.05 & 8.28 & 6.13 & \multicolumn{1}{c|}{13.56} & - & - & - & - & \multicolumn{1}{c|}{-} & \textbf{0.33} & \textbf{0.31} & 2.90 & \textbf{0.59} & 1.29\\
DECS & 7.25 & 9.23 & 18.84 & 13.05 & \multicolumn{1}{c|}{6.08} & 2.39 & 7.67 & 9.73 & 7.70 & \multicolumn{1}{c|}{6.11} & 7.25 & 9.87 & 14.55 & 16.78 & 14.22 \\ \midrule
BERT-M &  11.33          & 6.72 & 7.89 & 6.87 & \multicolumn{1}{c|}{3.47} & 1.77 & 1.77 & 2.12 & 1.77 & \multicolumn{1}{c|}{2.74} & 3.26 & 3.12 & 5.79 & 3.91 & 5.74 \\
BERT-E & 10.43 & 5.88 & 8.42 & 7.96 & \multicolumn{1}{c|}{4.23} & 1.67 & 1.56 & 2.41 & 2.41 & \multicolumn{1}{c|}{0.99} & 5.74 & 5.21 & 6.43 & 7.92 & 8.59 \\
RR & 6.72 & - & - & - & \multicolumn{1}{c|}{-} & 8.30 & - & - & - & \multicolumn{1}{c|}{-} & 7.44 & - & - & - & - \\
DPR & 5.05 & 3.14 & 2.89 & 1.94 & \multicolumn{1}{c|}{0.26} & 1.02 & 1.72 & 1.43 & 1.92 & \multicolumn{1}{c|}{2.25} & 2.09 & 0.86 & 0.66 & 9.26 & 6.88\\ \midrule
\textbf{Ours-H} & 3.80 & \textbf{2.36} & \textbf{2.42} & \textbf{1.77} & \multicolumn{1}{c|}{\textbf{0.36}} & \textbf{0.81} & \textbf{0.41} & \textbf{0.50} & \textbf{0.62} & \multicolumn{1}{c|}{\textbf{0.51}} & 0.81 & 0.41 & \textbf{0.50} &  0.62 & \textbf{0.51} \\ \bottomrule
\end{tabular}
\caption{The standard deviation of different methods for retrieving high-level components in terms of Acc, MRR, and top-$k$ hit ratios.
}\label{tab:std1}
\end{table*}

\begin{table*}[htb]
\centering
\footnotesize
\setlength\tabcolsep{2pt}
\begin{tabular}{@{}lccccccccccccccc@{}}
\toprule
                  & \multicolumn{5}{c|}{\textbf{SciREX}}                                                                         & \multicolumn{5}{c|}{\textbf{PubMed}}  & \multicolumn{5}{c}{\textbf{NLP-TDMS}}                                          \\ \midrule
\textbf{Methods}                   & \textbf{Acc}    & \textbf{MRR}    & \textbf{Hit@2}  & \textbf{Hit@3}  & \multicolumn{1}{c|}{\textbf{Hit@5}} & \textbf{Acc}    & \textbf{MRR}    & \textbf{Hit@2}  & \textbf{Hit@3}  & \multicolumn{1}{c|}{\textbf{Hit@5}} & \textbf{Acc} & \textbf{MRR} & \textbf{Hit@2} & \textbf{Hit@3} & \textbf{Hit@5} \\ \midrule
Base & 7.35 & 7.39 & 7.12 & 7.66 & \multicolumn{1}{c|}{11.50} & \textbf{0.57} & 1.02 & 0.82 & 0.84 & \multicolumn{1}{c|}{0.45} & 2.53 & 3.26 & 2.44 & \textbf{2.72} & \textbf{3.13} \\
SciREX & 13.87 & 9.63 & 13.28 & 11.45 & \multicolumn{1}{c|}{12.09} & 1.63 & 1.29 & 0.77 & 2.31 & \multicolumn{1}{c|}{1.87} & 3.95 & 3.60 & 3.35 & 4.92 & 4.83 \\ \midrule
GCN & 2.77 & \textbf{4.82} & \textbf{3.31} & 5.65 & \multicolumn{1}{c|}{8.60} & 1.59 & 1.02 & 2.06 & \textbf{0.75} & \multicolumn{1}{c|}{0.55} & 3.64 & 3.72 & 2.16 & 3.88 & 3.81 \\
GAT	& 3.32 & 4.89 & 4.17 & 6.17 & \multicolumn{1}{c|}{9.47} & 0.88 & \textbf{0.56} & 1.63 & 0.76 & \multicolumn{1}{c|}{\textbf{0.08}} & \textbf{2.38} & 2.99 & 2.97 & 3.11 & 3.19 \\
HDE	& 3.86 & 5.98 & 6.22 & 10.08 & \multicolumn{1}{c|}{13.07} & 1.88 & 1.04 & 2.43 & 2.87 & \multicolumn{1}{c|}{2.55} & 4.30 & 3.46 & 2.45 & 3.21 & 3.38 \\ \midrule
TAPAS &	\textbf{2.76} & - & - & - & \multicolumn{1}{c|}{-} & 1.07 & - & - & - & \multicolumn{1}{c|}{-} & 3.66 & - & - & - & - \\
TDMS-IE	& 4.39 & - & - & - & \multicolumn{1}{c|}{-} & 1.52 & - & - & - & \multicolumn{1}{c|}{-} & 4.87 & - & - & - & - \\ \midrule
Ours-L & 4.28 & 5.25 & 6.28 & \textbf{4.27} & \multicolumn{1}{c|}{0.70} & 1.12 & 1.91 & \textbf{0.21} & 0.80 & \multicolumn{1}{c|}{0.95} & 3.26 & \textbf{2.21} & \textbf{3.33} & 3.87 & 3.36 \\ \bottomrule
\end{tabular}
\caption{The standard deviation of different low-level methods in extracting the target entities.}\label{tab:std2}
\end{table*}

\begin{table*}[htb]\small
\centering
\vspace{-2ex}
\setlength\tabcolsep{2pt}
\begin{tabular}{@{}lccccccccccccccc@{}}
\toprule
                  & \multicolumn{5}{c|}{\textbf{SciREX}}                                                                         & \multicolumn{5}{c|}{\textbf{PubMed}}  & \multicolumn{5}{c}{\textbf{NLP-TDMS}}                                          \\ \midrule
\textbf{Methods}                   & \textbf{Acc}    & \textbf{MRR}    & \textbf{Hit@2}  & \textbf{Hit@3}  & \multicolumn{1}{c|}{\textbf{Hit@5}} & \textbf{Acc}    & \textbf{MRR}    & \textbf{Hit@2}  & \textbf{Hit@3}  & \multicolumn{1}{c|}{\textbf{Hit@5}} & \textbf{Acc} & \textbf{MRR} & \textbf{Hit@2} & \textbf{Hit@3} & \textbf{Hit@5} \\ \midrule
Base & 3.40 & \textbf{2.74} & \textbf{2.08} & \textbf{2.24} & \multicolumn{1}{c|}{3.65} & \textbf{1.35} & 2.14 & 3.72 & \textbf{1.79} & \multicolumn{1}{c|}{\textbf{2.59}} & \textbf{0.73} & \textbf{0.16} & \textbf{0.07} & \textbf{0.43} & \textbf{1.15} \\
GCN	& 3.19 & 3.76 & 3.08 & 3.08 & \multicolumn{1}{c|}{2.70} & 3.16 & \textbf{1.94} & 2.41 & 3.30 & \multicolumn{1}{c|}{3.29} & 0.80 & 0.56 & 0.74 & 1.41 & 1.18 \\
GAT	& \textbf{2.12 }& 3.27 & 2.62 & 8.19 & \multicolumn{1}{c|}{\textbf{1.80}} & 5.85 & 2.05 & 5.73 & 3.83 & \multicolumn{1}{c|}{6.52} & 1.49 & 1.23 & 0.57 & 1.63 & 2.09 \\
BERT-E+B & 8.58 & 6.36 & 3.11 & 6.04 & \multicolumn{1}{c|}{2.73} & 3.41 & 6.73 & 5.82 & 5.01 & \multicolumn{1}{c|}{6.16} & 1.28 & 2.22 & 1.44 & 1.25 & 1.38 \\
Ours & 3.21 & 3.45 & 2.29 & 4.85 & \multicolumn{1}{c|}{2.46} & 2.86 & 3.77 & \textbf{1.45} & 3.54 & \multicolumn{1}{c|}{3.66} & 1.43 & 1.10 & 1.69 & 0.94 & 1.94 \\ \bottomrule
\end{tabular}
\caption{The standard deviation of overall document-level extraction performance of different methods. 
}\label{tab:std3}
\end{table*}

%% file: appendices/appendix3.tex
\section{Implementation Details}
\label{appendix:implementation}
\subsection{Hyper-parameters Settings}\label{subapp:hyper}
For the high-level component retriever training, the learning rate is set as $1e-4$ and the maximum number of epochs is $50$. 
For low-level entity selector training, we use a 1-layer single head GAT~\cite{velivckovic2018graph} based on the bag-of-neighbors features computed on 1-st order neighbors to aggregate graph topology information. 
We select $\lambda=0.3$ and $\mu=0.15$ as the proportion weights in the multi-view aggregation.
For the feature-based FFNN classifiers in both high-level and low-level models, we set the dimensions of the hidden layers to $32$.
The corresponding learning rate and maximum number of epochs to the low-level entity extractor are $1e-3$ and $50$. 
During training, we use the Adam~\cite{kingma2014adam} optimizer with $\beta_1=0.9$ and $\beta_2=0.999$ in our experiments for all the models.
We select the best set of hyper-parameters of the models based on the accuracy on the corresponding dev sets.

\subsection{Implementation Settings}\label{subapp:imple}
We train and test our code on the System Ubuntu 18.04.4 LTS with CPU: Intel(R) Xeon(R) Silver 4214 CPU@ 2.20GHz and GPU: NVIDIA GeForce RTX 2080.
We implement our method using Python 3.8 and PyTorch 1.6~\cite{paszke2019pytorch}.

\section{Dataset Description}
\label{appendix:dataset}
We evaluate our work on three different datasets (see Table \ref{tab:data}): 
(1) \textbf{SciREX}~\cite{jain2020scirex}  contains 438 annotated full-length papers, related to machine learning research. We extend the original SciREX dataset by extracting the tabular data of each paper from the corresponding raw LaTeX or PDF files. For this dataset, the queries are in the format of \texttt{<Task, Method, Dataset, Metric>} and the final answer we aim to find from the documents is the corresponding \texttt{score}; 
(2) \textbf{NLP-TDMS (Full)}~\cite{hou2019identification} contains 332 unannotated full-length papers, including both the text data and the tabular data, related to the natural language processing research. For this dataset, the queries are in the format of \texttt{<Task, Dataset, Metric>} and we are looking for the corresponding \texttt{score}s; 
(3) \textbf{PubMed}~\cite{jia2019document} is created by automatically
labeling biomedical literature with Gene Drug Knowledge Database. The dataset
contains 5688 annotated full-length papers, related to biochemical research.
The queries designed for this dataset are in the format of \texttt{<Gene,
  Drug>}, and the task is to extract  the most influenced \texttt{Mutations}.

\section{Detailed Evaluation Protocol}\label{appendix:eval}
Given $n$ samples in the test set, assume that $\{\hat{y}_1,\cdots,\hat{y}_n\}$ and $\{y_1,\cdots,y_n\}$ are the model predictions and ground-truth labels, respectively. Besides, $\{\mathbf{\hat{y}}_1^{k},\cdots,\mathbf{\hat{y}}_n^k\}$ indicates the top-$k$ selections made by the models for each test example. The high-level and low-level models use the same set of evaluation metrics, with the only difference that the high-level models use component-level labels, while the low-level models use the entity-level labels. We use the following metrics for all the methods:
(1)\textbf{Accuracy (Acc)} measures the exact match for querys in the test set. It only counts the cases when the prediction equals to the ground truth: $$\operatorname{Acc}=\frac{1}{n}\sum_{i=1}^n\mathbb{I}(y_i=\hat{y}_i);$$
(2) \textbf{Mean Reciprocal Rank (MRR)} is the average reciprocal ranks of a query's ground-truth answer among all the candidates: $$\operatorname{MRR}=\frac{1}{n}\sum_{i=1}^n\frac{1}{rank(y_i)};$$
(3) \textbf{Top-k Hit Rate (Hit@K)} measures whether the ground-truth answer is included in the top-k selection made by the models: $$\operatorname{Hit@k}=\frac{1}{n}\sum_{i=1}^n\mathbb{I}(y_i\in\mathbf{\hat{y}}_i^k).$$
For Top-k Hit Rate, we report Hit@2, Hit@3, and Hit@5. For the high-level scenario, we only evaluate whether the model select the correct component or not; For the low-level scenario, we evaluate the performances restricting the searching space into the ground-truth paragraph/table for entity selection; For the overall scenario, we remove the restriction and test the model performance of entity selection from the whole document.

%% file: appendices/appendix4.tex
\section{Experimental Baselines}\label{appendix:baseline}
We use different baselines for the high-level component retrieval, low-level entity selection, and the overall framework. 

\paragraph{High-Level Baselines.} For the high-level model, we compare with the following baselines:

\noindent$\bullet~$\textbf{Sparse Retrieval Methods:} 1) \textbf{TF-IDF}~\cite{aizawa2003information} and 2) \textbf{BM25}~\cite{robertson2009probabilistic}  are two sparse-retrieval methods which ranks query-section pairs via computing the relevant score based on key words;

\noindent$\bullet~$\textbf{Entity-Based Methods:} 1) \textbf{Entity Cosine Similarities (ECS)}  calculates the cosine similarities between the embeddings of query and section entities, and sums them up as the final prediction score; 2) \textbf{Deep Entity Cosine Similarities (DECS)} improves cosine similarities by substituting the sum-up function into a feedforward neural network.

\noindent$\bullet~$\textbf{Embedding-Based Methods:} 1) \textbf{BERT-Matching}
is a matching method based on pre-trained BERT embeddings, using the dot product between query and component representations; 2) \textbf{BERT-Entailment} is a textual inference method
~\cite{nogueira2019passage, nie2019revealing} for calculating the relevance score.
3) \textbf{Recurrent
  Retriever}~\cite{asai2019learning} is a graph-based recurrent retrieval
method. 
It selects one paragraph $p_i$ in each step until it selects an
end-of-evidence mark ([EOE]); 
4) \textbf{Dense Passage Retrieval
  (DPR)}~\cite{karpukhin2020dense} is a state-of-the-art model that use BERT as the encoder for passage retrieval in open-domain QA.

\paragraph{Low-Level Baselines.} For the low-level model, we restrict the searching space to the ground-truth paragraph/table that contains the final answer and compare with the following baselines:

\noindent$\bullet~$\textbf{Embedding-Based Methods:} 1) \textbf{BERT-Base} is a simple classifier trained directly on the
concatenation of query and candidate embeddings; 2)
\textbf{SciREX}~\cite{jain2020scirex} composes salient entity embeddings for
each paragraph and learns a binary classifier to decide whether the $N$-ary relation exists or not.

\noindent$\bullet~$\textbf{Graph-Based Methods:} 1) \textbf{Graph Convolutional Network (GCN)}~\cite{kipf2017semi} and 2) \textbf{Graph Attention Network (GAT)}~\cite{velivckovic2018graph} are two classic graph neural network structures, we report performances by applying them on our proposed graph structure; 3) \textbf{Heterogeneous Document-Entity (HDE) graph}~\cite{tu2019multi} is a heterogeneous graph model which conducts  multi-hop reading comprehension by leveraging the relation between document, entity, and candidate nodes;

\noindent$\bullet~$\textbf{Pre-trained LMs:} 1) \textbf{TAPAS}~\cite{herzig2020tapas} is the start-of-the-art pre-trained model on text and tables. 
We fine-tune the pre-trained model on our own datasets; 2) \textbf{TDMS-IE}~\cite{hou2019identification} is an entailment model based on the score context and hypothesis of dataset and metric to judge whether these elements are related to each other.

\paragraph{Overall Baselines.} For the overall performance our two-stage model, we compare with the following baselines: 1) The \textbf{BERT-Base} model searching in the whole document; 2) \textbf{GCN} and 3) \textbf{GAT} testing the performance of our proposed graph in the whole document; 4) \textbf{BERT-Entailment+Base} is a two-stage model combining the best baselines for both high-level and low-level stage.

%% file: appendices/appendix5.tex
\section{Standard Deviation of Main Results}\label{appendix:std}
Table~\ref{tab:std1}--Table~\ref{tab:std3} list the standard deviations we obtain for the main results from multiple trials. 
The results indicate that \ours show competitive stability compared with all the baselines on three different datasets under different settings.
The evaluation computation is based on the number of queries, but we split the training, validation, and test set based on the number of documents to prevent data leakage. 
Due to the fact that various documents include varying numbers of queries, the exact number of queries in train/val/test set is not fixed, causing the performances to vary in different trials and the standard deviations to increase. 
For PubMed dataset, the test set is fixed and the random seeds can only influence the split between training and validation set, thus the standard deviations on this dataset is relatively smaller than on the other two datasets, SciREX and NLP-TDMS.